\documentclass[sigconf]{acmart}
\usepackage{subfigure}
\usepackage{amsmath}
\usepackage{bm}
\AtBeginDocument{%
  }

\copyrightyear{2025}
\acmYear{2025}
\setcopyright{acmlicensed}
\acmConference[MM '25]{Proceedings of the 33rd ACM International Conference on Multimedia}{October 27--31, 2025}{Dublin, Ireland}
\acmBooktitle{Proceedings of the 33rd ACM International Conference on Multimedia (MM '25), October 27--31, 2025, Dublin, Ireland} \acmDOI{10.1145/3746027.3754933}
\acmISBN{979-8-4007-2035-2/2025/10}
\begin{document}

\title{MEDTalk: Multimodal Controlled 3D Facial Animation with Dynamic Emotions by Disentangled Embedding}

\author{Chang Liu}
\email{frenkiedejong@sjtu.edu.com}
\affiliation{%
  \institution{Shanghai Jiao Tong University}
  \city{Shanghai}
  \country{China}
}

\author{Ye Pan}
\authornote{Corresponding author}
\email{whitneypanye@sjtu.edu.cn}
\affiliation{%
  \institution{Shanghai Jiao Tong University}
  \city{Shanghai}
  \country{China}
}

\author{Chenyang Ding}
\email{chenyangding@sjtu.edu.com}
\affiliation{%
  \institution{Shanghai Jiao Tong University}
  \city{Shanghai}
  \country{China}
}

\author{Susanto Rahardja}
\email{susantorahardja@ieee.org}
\affiliation{%
  \institution{Singapore Institute of Technology}
  \country{Singapore}
}

\author{Xiaokang Yang}
\email{xkyang@sjtu.edu.com}
\affiliation{%
  \institution{Shanghai Jiao Tong University}
  \city{Shanghai}
  \country{China}
}

\begin{abstract}
Audio-driven emotional 3D facial animation aims to generate synchronized lip movements and vivid facial expressions. However, most existing approaches focus on static and predefined  emotion labels, limiting their diversity and naturalness. To address these challenges, we propose \textbf{MEDTalk}, a novel framework for fine-grained and dynamic emotional talking head generation. Our approach first disentangles content and emotion embedding spaces from motion sequences using a carefully designed cross-reconstruction process, enabling independent control over lip movements and facial expressions. Beyond conventional audio-driven lip synchronization, we integrate audio and speech text, predicting frame-wise intensity variations and dynamically adjusting static emotion features to generate realistic emotional expressions. Furthermore, to enhance control and personalization, we incorporate multimodal inputs—including text descriptions and reference expression images—to guide the generation of user-specified facial expressions. With MetaHuman as the priority, our generated results can be conveniently integrated into the industrial production pipeline. The code is available at: \href{https://github.com/SJTU-Lucy/MEDTalk}{https://github.com/SJTU-Lucy/MEDTalk}.

\end{abstract}





\maketitle

\section{Introduction}
3D talking head generation aims to synthesize realistic facial animations from speech or text inputs and has become a significant research area in computer graphics and artificial intelligence. This technology has broad applications in virtual assistants, digital avatars, gaming, and education \cite{korban2022survey, al2021review}, enhancing human-computer interaction through lifelike digital characters. Compared to traditional manual animation, which is time-consuming and labor-intensive, data-driven approaches significantly improve efficiency while producing expressive and natural facial animations.

\begin{figure}[t]
  \centering
  \includegraphics[width=\linewidth]{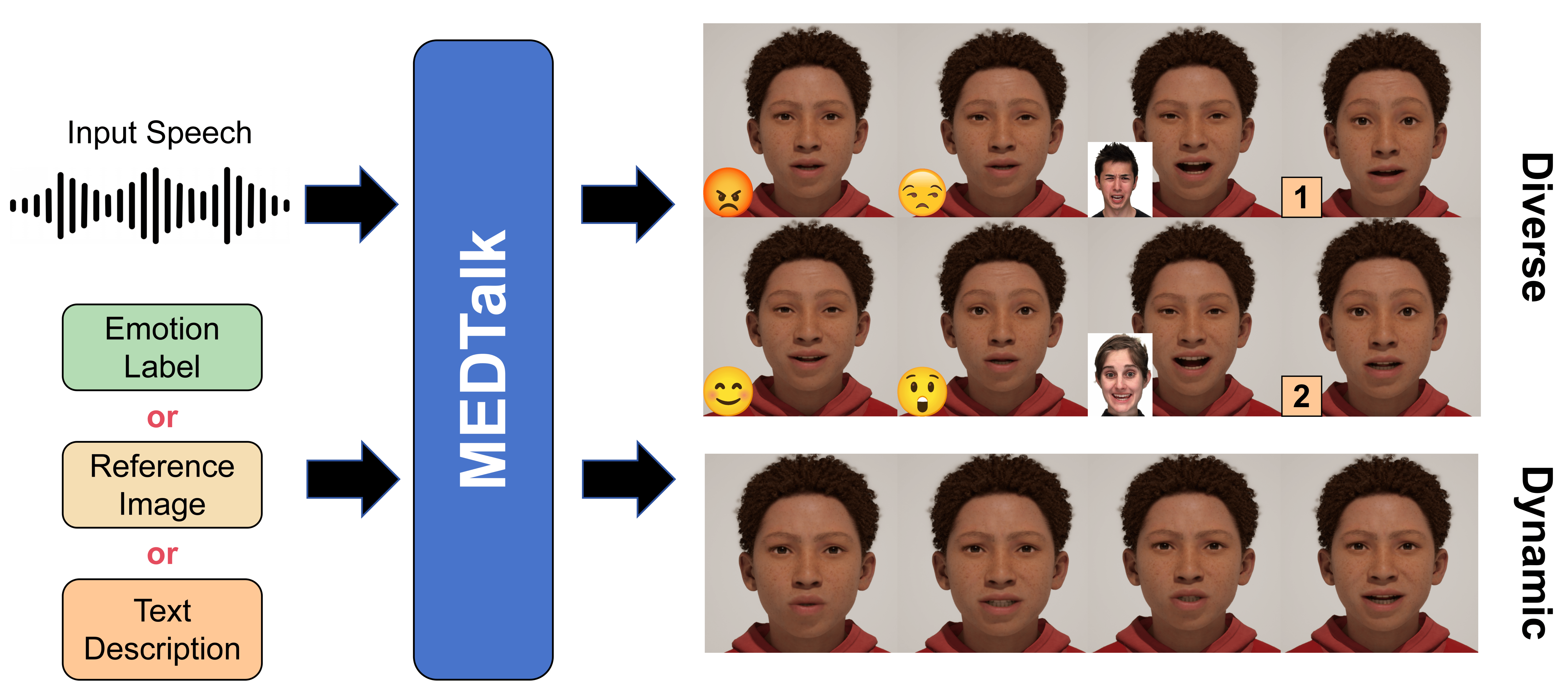}
  \caption{Results of MEDTalk. Given an input speech, MEDTalk produces expressive facial movements based on emotion labels, reference images, or text descriptions, with natural emotional transitions.} 
  \label{fig:teaser}
\end{figure}

Early studies \cite{edwards2016jali, pham2017speech, pham2018end, zhou2018visemenet, cudeiro2019capture, richard2021meshtalk} primarily focused on lip synchronization while neglecting movements in other facial regions, leading to unnatural animations. Recent methods have introduced emotion control by incorporating emotion or intensity labels \cite{fan2022faceformer, pan2023emotional, liu2024emoface}, enhancing the expressiveness of facial animations. However, label-based approaches impose a fixed emotional state throughout the animation, limiting diversity and expressiveness. Some studies \cite{peng2023emotalk, kim2024deeptalk} try to extract emotion features from the input audio to improve the diversity of facial expressions. However, emotional features are not fully represented in audio, reflected in inaccurate predictions. We argue that high-quality facial animation relies on two key components: lip synchronization and emotional expressiveness. Lip movements must align precisely with speech audio, while emotional expressions should exhibit both diversity and dynamic micro-expressions to enhance realism.

To address these issues, we propose \textbf{MEDTalk}: \textbf{M}ultimodal Controlled 3D Facial Animation with Dynamic \textbf{E}motions by \textbf{D}isentang-led Embedding, as shown in Fig. \ref{fig:teaser}. Our framework disentangles emotion and content features through self-supervised training on motion sequences, enabling independent control of lip synchronization and facial expressions. Apart from precise control over lip synchronization, MEDTalk also supports multiple input modalities, including conventional emotion labels, textual descriptions, and reference images, allowing for fully controllable expression generation. Additionally, we employ cross-attention to integrate features of audio and spoken text, predicting frame-wise emotion intensity to get dynamic emotion features. This approach enables the generation of highly accurate lip-synced facial animations with fine-grained, controllable, and dynamically evolving emotions.

Our key contributions are as follows:
\begin{itemize}
    \item We introduce MEDTalk, an emotional 3D talking head framework that generates facial animation with fine-grained and multimodal-guided emotion states.
    \item We propose a novel paradigm that disentangles content and emotion through self-supervised training, enabling independent control of lip and facial expressions.
    \item We integrate audio and speech text to predict frame-wise emotional intensity without requiring frame-wise annotations, generating dynamic emotional facial animation.
    \item We introduce multimodal guidance, overcoming the limitations of predefined labels in previous studies and enabling user control, improving diversity in facial animation.
\end{itemize}

\section{Related Work}
\subsection{Audio-driven 3D Facial Animation}
The goal of audio-driven talking head generation is to animate a target character based on an audio clip while ensuring accurate lip synchronization. With the advancement of deep learning, various model architectures have been utilized to generate expressive and realistic 3D facial animations.

Early works \cite{pham2018end, cudeiro2019capture} primarily focused on obtaining realistic, speaker-independent lip shapes from mel spectrograms using convolutional neural networks (CNNs). Transformer-based approaches, such as FaceFormer \cite{fan2022faceformer}, leveraged wav2vec2.0 \cite{baevski2020wav2vec} for audio feature extraction and synthesized continuous motion via autoregressive modeling, improving lip synchronization accuracy.
CodeTalker \cite{xing2023codetalker} employed a VQ-VAE framework, treating facial animation generation as a discrete code query task, enhancing robustness.
VividTalker \cite{zhao2023breathing} further refined this approach by encoding head pose and mouth motion into separate discrete latent spaces, using audio features for autoregressive motion prediction.
Meanwhile, SelfTalk \cite{peng2023selftalk} introduced a self-supervised framework that aligns generated lip shapes with corresponding speech text, incorporating text-audio consistency as additional supervision.
Recent methods \cite{chen2023diffusiontalker, stan2023facediffuser, sun2024diffposetalk} have explored the diffusion model \cite{ho2020denoising}, leveraging its generative diversity to enhance realism and motion diversity in talking head generation

Despite these advancements, most existing methods primarily focus on audio-synchronized lip movements while neglecting expressive variations in other facial regions. As a result, generated animations often appear rigid and emotionless. In contrast, our approach, MEDTalk, allows for expressive and natural facial animations with fine-grained emotional variations.

\subsection{Emotional Talking Head}
Emotional talking head generation incorporates emotional information, either manually controlled or extracted from audio, to produce facial animations with expressive emotional features.

Some works apply predefined emotion labels to control the generated facial expression, such as EMOTE \cite{danvevcek2023emotional}, EmoFace \cite{liu2024emoface}, and FlowVQTalker \cite{tan2024flowvqtalker}.
Some works have explored extracting emotion features from audio to improve expressiveness \cite{ji2021audio, peng2023emotalk, tan2023emmn}. They applied cross-reconstruction to separate emotional cues from speech. 
DEEPTalk \cite{kim2024deeptalk} constructed a joint feature space for audio and expressions, connecting the emotional features between them.
AVI-Talking \cite{sun2024avi} utilized a large language model (LLM) to analyze emotion features in speech.

Despite these advancements, most existing methods rely on static emotion features, resulting in facial animations that lack variation. To address this, DEITalk \cite{shen2024deitalk} combined emotion labels with dynamically estimated emotion intensities derived from a speech-expression joint space. However, the predicted intensity lacked supervision and was heavily correlated with input audio. In contrast, our approach leverages both audio and textual modalities to predict dynamic emotional intensity in a supervised manner, enhancing the stability of generated emotional variations.

\subsection{Multimodal Guided Generation}
Integrating multimodal information into generative models remains a key focus in Artificial Intelligence-Generated Content (AIGC). In talking head generation, combining diverse data sources such as text and visual cues enables realistic and expressive facial animations.

Aligned Multi-modal Emotion encoder \cite{xu2023high} embeds text, image, and audio in a unified space. Leveraging the rich semantic priors of CLIP, this method allows for multimodal guidance in facial expression generation. 
EAT \cite{gan2023efficient} proposed an adaptive Emotional Deformation Network, which generates emotion-compliant talking heads based on either emotional labels or text descriptions.

Recent studies  \cite{ma2023talkclip, zhong2024expclip, zhao2024media2face, tan2024style2talker, tan2024say, tan2025edtalk, tan2025fixtalk} have explored one-shot talking head generation using text descriptions or reference expression images. By extracting the style of multimodal guidance through a carefully designed feature encoder, these methods utilized textual or visual inputs indicating desired speaking styles, enabling realistic talking head synthesis without requiring reference videos. 
PMMTalk \cite{han2024pmmtalk} integrates an off-the-shelf talking head generation method with speech recognition, extracting visual and textual cues from speech. It aligns audio, image, and text features at the temporal and semantic levels to guide generation. 
MMHead \cite{wu2024mmhead} extends this approach to text-guided 3D head animation, supporting text-to-3D facial motion generation by hierarchical text annotation.

Building on these advancements, our method enables text descriptions or reference images to guide synthesis, generating nuanced expressions beyond predefined emotion labels.

\begin{figure*}[h]
  \centering
 \includegraphics[width=\linewidth]{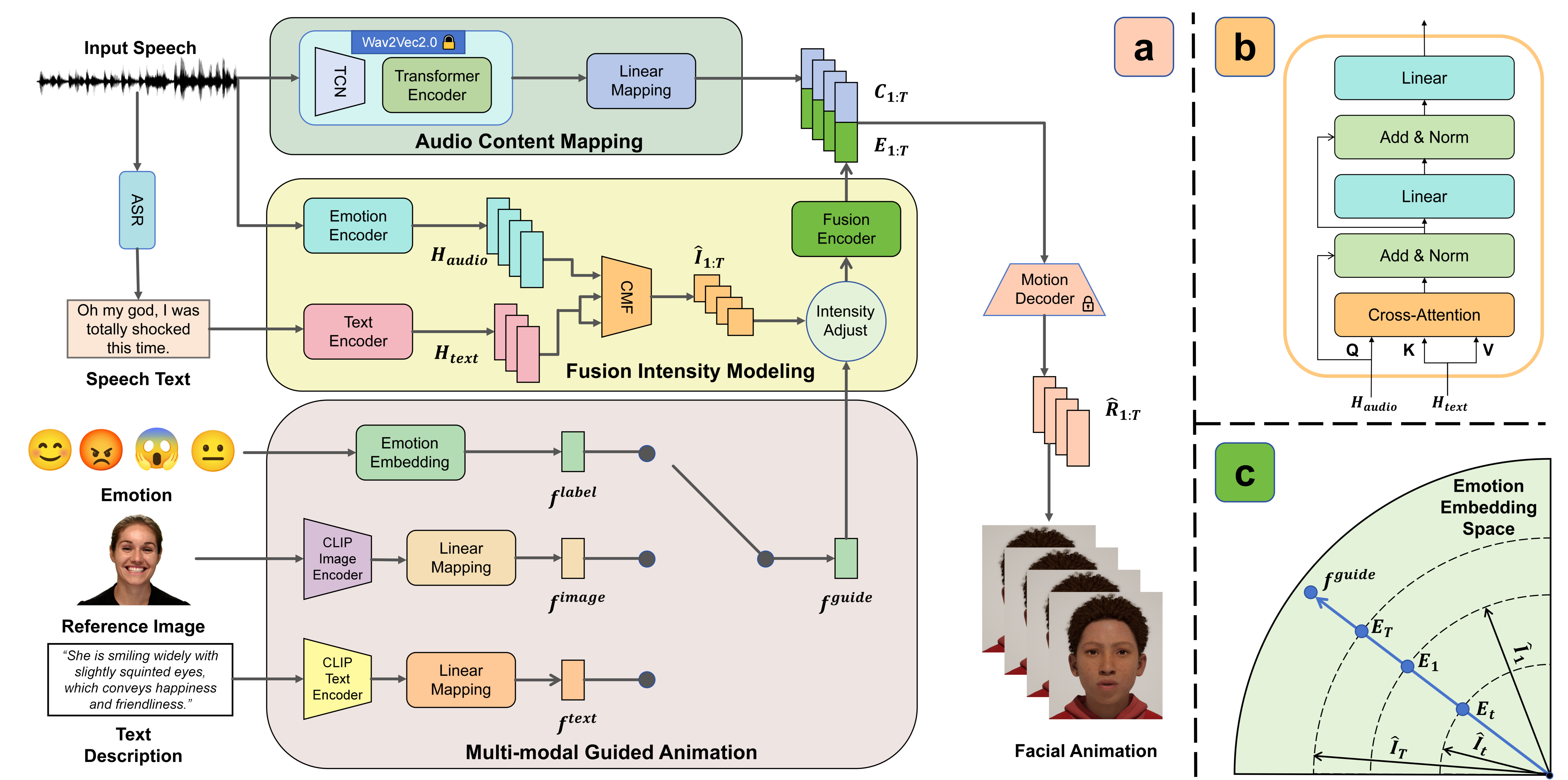}
    \caption{Illustration of MEDTalk. (a) The overall architecture of MEDTalk. It integrates audio, text, and multimodal guidance to generate expressive facial animations. (b) The Cross-Modality Fusion (CMF) module, which effectively combines audio and text features. (c) Intensity adjustment. The norm of emotion embedding is modulated based on the predicted emotion intensity.}
    \label{fig:pipeline}
\end{figure*}

\section{Methods}
\subsection{Overview}

We proposed MEDTalk, which generates facial animations with dynamic and diverse emotions guided by multimodal inputs and accurate synchronization with audio. The overall structure is illustrated in Fig. \ref{fig:pipeline}. Our novel generation architecture consists of three key components: disentangled content and emotional latent space, audio-driven dynamic emotional facial expression generation, and multimodal guided stylized expression generation. By integrating these components, MEDTalk achieves separate control over lip shape and emotional expression, while supporting precise facial animation guided by multimodal inputs.

Let $\bm{A}_{1:T} = (a_1, ..., a_T)$ be a sequence of input speech snippets with each $a_t \in \mathbb{R}^{D}$ containing D sampled audio, and let $R_{1:T}=(r_1,...,r_T)$, where $r_t\in \mathbb{R}^{174}$ is a sequence of MetaHuman controller rig. For emotion guidance, let $\bm{G}\in \{g^{label},g^{image},g^{text} \}$ be the user-controllable guidance, which can be either an emotion label, a reference image, or a text description. Our goal is to take the audio and emotion guidance as inputs and predict the emotional facial rig sequence $\hat{\bm{R}} _{1:T}=(\hat{r}_1,...,\hat{r}_T)$ that aligns with the input speech. The end-to-end procedure can be written as:

\begin{equation}
\hat{\bm{R}} _{1:T}=MEDTalk(\bm{A}_{1:T}, \bm{G}) 
\end{equation}

While multiple modalities can be used for emotion guidance, other model components remain shared across different modalities, ensuring consistency and efficiency in generation.

\subsection{Disentangled Embedding Space} \label{sec:disentangle}

Some prior studies \cite{ji2021audio, peng2023emotalk, nocentini2024emovoca} have attempted to separate content and emotion features from speech signals, allowing direct emotion extraction from speech for facial animation generation. However, since emotion representation in speech is inherently incomplete, even state-of-the-art speech emotion recognition (SER) models \cite{chen2022wavlm, ma2023emotion2vec, tang2023salmonn, radford2023robust, chen2024emova, team2024gemini, fu2025vita} fail to achieve fully accurate predictions. This uncertainty makes emotion extraction from speech unreliable for precise avatar generation applications.

Therefore, we adopt a self-reconstruction approach using facial motion, rather than speech, to separate content and emotion features. Compared to audio-based disentanglement, our method offers three key advantages:
\begin{enumerate}
    \item Facial motion directly expresses emotional expressions, avoiding the need for complex many-to-many audio-expression mappings. The effectiveness of the separation can be directly validated by reconstruction.
    \item Cross-training requires perfectly aligned sequences of equal length with the same content but different emotions. In audio, collecting such data is challenging, and Dynamic Time Warping (DTW) \cite{muller2007dynamic} is required for sequence alignment, which could introduce distortions. In contrast, we can generate aligned sequences by applying identical speech content with different emotion labels.
    \item After training, freezing the motion decoder allows features from different modalities to be mapped into a unified embedding space. This simplifies multimodal control and ensures consistent emotion representation across modalities.
\end{enumerate}

\begin{figure}[t]
  \centering
  \includegraphics[width=\linewidth]{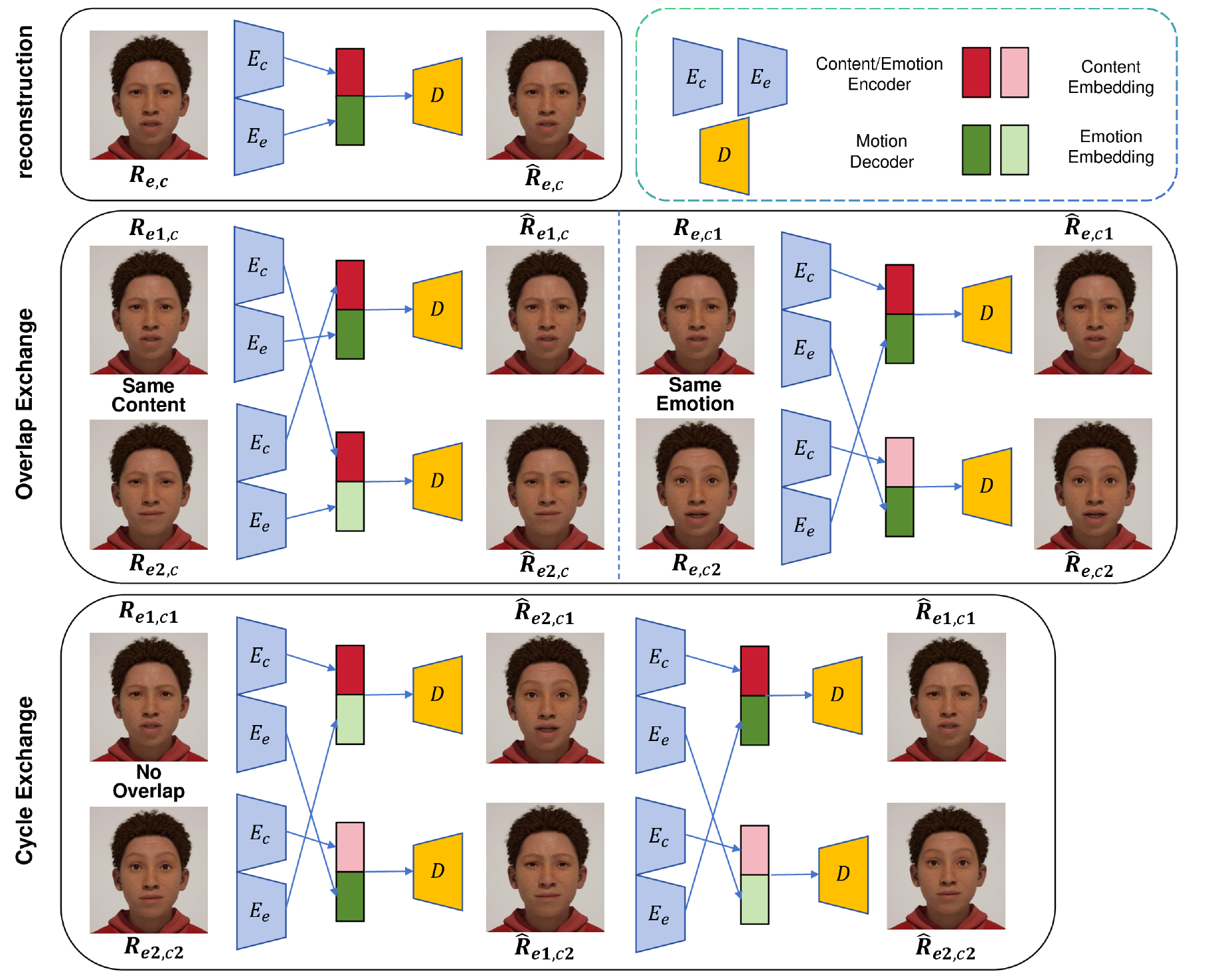}
  \caption{Illustration of our disentanglement strategy using self-supervised learning, which can effectively decouple content and emotion features.}
  \label{fig:disentangle}
\end{figure}

To obtain data for cross-reconstruction training, we leverage EmoFace \cite{liu2024emoface}, a pre-trained model that takes audio and emotion labels as inputs and predicts facial animation. Specifically, we generate speech samples with seven emotions to create a large-scale dataset. It enables flexible content-emotion pairing during training.

Through our experiments, we found that mutual information minimization, such as Gradient Reversal Layer (GRL) \cite{ganin2015unsupervised} and vCLUB \cite{cheng2020club} used in voice conversion tasks \cite{yang2022speech, williams2021learning}, was insufficient for effective content-emotion separation, leading to suboptimal generation quality. In contrast, we draw inspiration from \cite{ge2020zero} and adopt a novel training scheme, as illustrated in Fig. \ref{fig:disentangle}, achieving high-quality feature disentanglement. 

Specifically, our decoupling network employs two encoders, $E_e$ and $E_c$, to extract emotion and content features from the input motion sequence. These features are concatenated and passed through a decoder $D$, which reconstructs the facial animation sequence. Let $\bm{R}_{e,c}$ denote the input motion sequence with emotion $e$ and content $c$. To enforce disentanglement, we introduce a self-supervised feature exchange strategy, consisting of three key phases:

\begin{enumerate}
    \item \textbf{Self-reconstruction}: Ensures the accuracy of the reconstructed motion sequence. We reconstruct the motion itself:
    \begin{equation}
        \hat{\bm{R}}_{e,c}=D(E_e(\bm{R}_{e,c}),E_c(\bm{R}_{e,c}))
    \end{equation}
    
    \item \textbf{Overlap exchange}: Encourages embeddings of the same attribute to be close in feature space. Given two sequences with the same content (or emotion), we swap the corresponding feature representations:
    \begin{equation}
        \begin{cases}
            \hat{\bm{R}}_{e1,c}=D(E_e(\bm{R}_{e1,c}),E_c(\bm{R}_{e2,c})) \\
            \hat{\bm{R}}_{e2,c}=D(E_e(\bm{R}_{e2,c}),E_c(\bm{R}_{e1,c}))
        \end{cases}
    \end{equation}

    \item \textbf{Cycle exchange}: Prevents feature loss by ensuring consistency through repeated cross-reconstruction. Given an arbitrary set of motion sequences, we perform two rounds of cross-reconstruction:
    \begin{equation}
        \begin{cases}
        \hat{\bm{R}}_{e1,c2}=D(E_e(\bm{R}_{e1,c1}),E_c(\bm{R}_{e2,c2})) \\
        \hat{\bm{R}}_{e2,c1}=D(E_e(\bm{R}_{e2,c2}),E_c(\bm{R}_{e1,c1})) \\
        \hat{\bm{R}}_{e1,c1}=D(E_e(\hat{\bm{R}}_{e1,c2}), E_c(\hat{\bm{R}}_{e2,c1})) \\
        \hat{\bm{R}}_{e2,c2}=D(E_e(\hat{\bm{R}}_{e2,c1}), E_c(\hat{\bm{R}}_{e1,c2}))
        \end{cases}
    \end{equation}
\end{enumerate}

Training is supervised solely by the reconstruction loss $\mathcal{L}_{recon}$, which is the $\ell_2$ error of the reconstructed sequences.

Once the content-emotion embedding spaces are effectively disentangled, we freeze the encoders and the decoder. This enables seamless mapping of features from other modalities into the corresponding embedding space, ensuring flexible and accurate facial animation generation.

\subsection{Dynamic Emotional Facial Animation} \label{sec:dynamic_emotion}

In this section, we introduce an approach for generating emotional facial animations with dynamic intensity using audio and emotion labels as inputs. The model is illustrated in Fig. \ref{fig:pipeline}(a) and comprises two primary modules: Audio Content Mapping (ACM) and Fusion Intensity Modeling (FIM), responsible for mapping content and emotion features, respectively. The extracted features are concatenated and fed into the frozen decoder, generating dynamic emotional facial animations. Formally, the process can be expressed as:

\begin{equation}
    \hat{\bm{R}}_{1:T}=D(FIM(\bm{A}_{1:T}, g^{label}),ACM(\bm{A}_{1:T}))
\end{equation}

\paragraph{\textbf{Audio Content Mapping}} Prior studies \cite{peng2023emotalk, liu2024emoface} have demonstrated that wav2vec2.0 \cite{baevski2020wav2vec} effectively extracts lip-related features from input audio $\bm{A}_{1:T}$. Consequently, we extract audio content features via wav2vec2.0 and a lightweight mapping network. Our approach achieves perfect lip-sync with fast inference, making it practical for real-time applications.

During training, the predicted audio content embeddings, combined with the emotion embeddings generated by the frozen emotion encoder, serve as inputs to the frozen decoder, which predicts facial expressions. In addition to the reconstruction loss $\mathcal{L}_{recon}$, we introduce a similarity loss $\mathcal{L}_{sim}$, which measures the cosine similarity of content embedding from the audio and motion:

\begin{equation}
    \mathcal{L}_{sim}=1-cos(E_c(\bm{R}_{1:T}),ACM(\bm{A}_{1:T}))
\end{equation}

\begin{equation}
    \mathcal{L}_{ACM}=\mathcal{L}_{recon}+\lambda_{sim}\mathcal{L}_{sim}
\end{equation}

\paragraph{\textbf{Fusion Intensity Modeling}} We utilize seven emotional labels to control the emotions of the generated expressions. However, it lacks dynamic emotional changes in the generated expression as a single label guides the entire clip. To address this limitation, we integrate audio and speech text features to predict frame-wise emotional intensity, thereby enabling the generation of emotional expressions with dynamic intensity.

\begin{figure}[t]
  \centering
  \includegraphics[width=0.9\linewidth]{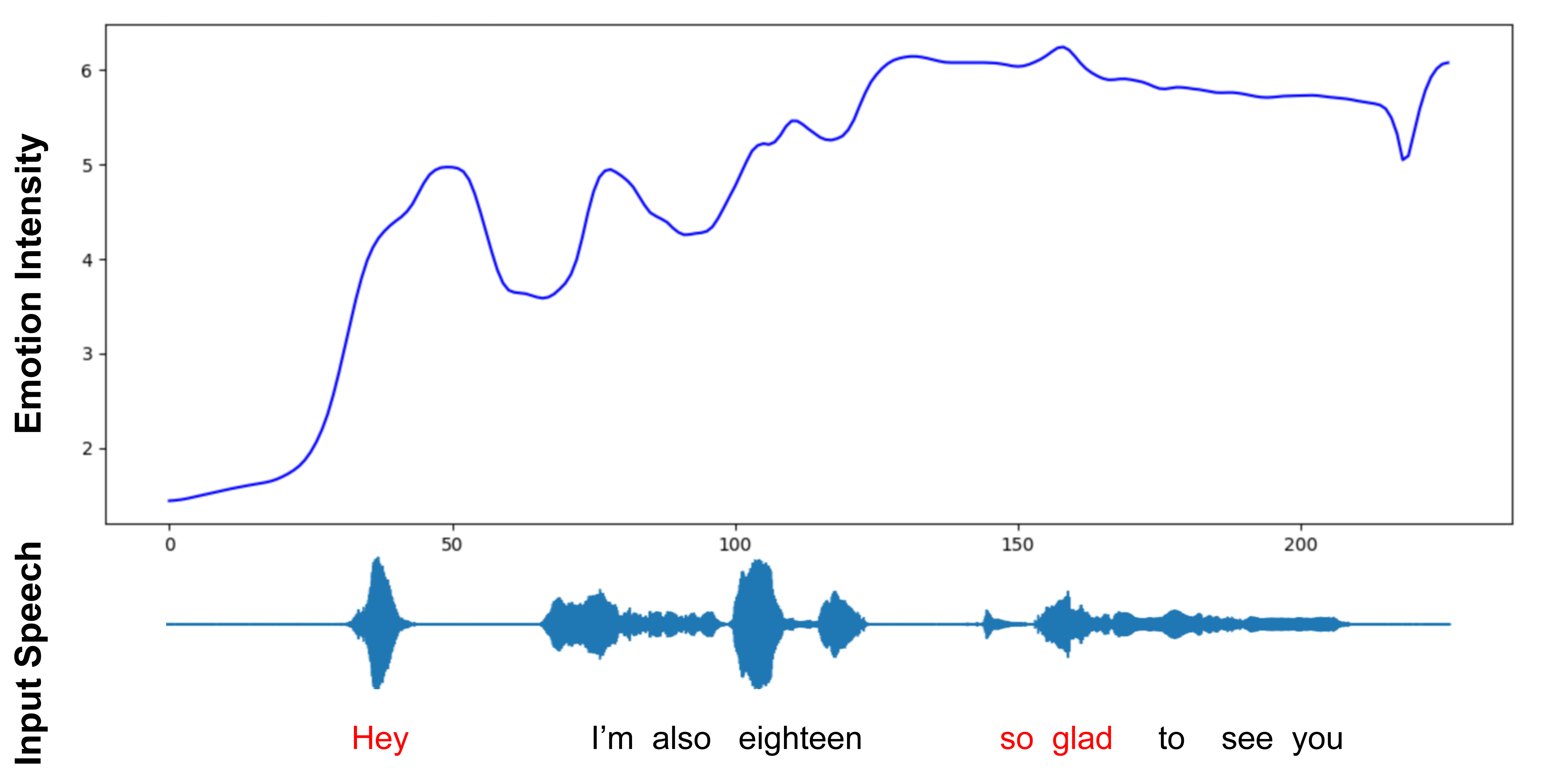}
  \caption{Visualization of intensity fluctuation with audio, where intensity increases when words with strong emotional semantics are detected.} 
  \label{fig:intensity}
\end{figure}

Since manually annotating frame-wise emotional intensity is impractical, we derive intensity labels from the motion sequence. Specifically, emotional intensity is primarily reflected in the upper face and overall lip shape. We select specific controller rigs and compute their L1-norm as the pseudo-intensity.

\begin{equation}
Int(\bm{R}_{1:T}) = \left( \Vert r_{1}^{S_{int} }  \Vert_1 ,..., \Vert r_{T}^{S_{int} }  \Vert_1 \right)
\end{equation}

where $S_{int}$ is the set of controllers selected to represent emotion intensity. Fig. \ref{fig:intensity} illustrates how the intensity fluctuates with the audio and speech text.

To predict frame-wise emotional intensity, we leverage both prosodic and semantic information. Precisely, we extract frame-wise emotion-related features $\bm{H}_{audio}$ from audio using emotion2vec \cite{ma2023emotion2vec}, avoiding explicit emotion recognition. As for speech text, we utilize Whisper \cite{radford2023robust} to transcribe audio into text and encode using a pre-trained RoBERTa-based model \cite{cui2020revisiting}, obtaining text embeddings $\bm{H}_{text}$. These two modalities are integrated through the Cross-Modality Fusion (CMF) module, which employs cross-attention to predict frame-wise intensity $\hat{\bm{I}}_{1:T}$, as depicted in Fig. \ref{fig:pipeline}(b).

The emotion embedding $f^{label}$ is adjusted using the predicted frame-wise intensity. To ensure that emotion features remain discretely distributed and their spatial location reflects emotional intensity, we define the norm of the emotion feature as the emotion intensity. We obtain dynamic emotion features $\tilde{\bm{F}}^{label}_{1:T}=(\tilde{f}^{label}_1, ..., \tilde{f}^{label}_T)$ by adjusting the norm frame by frame, as illustrated in Fig. \ref{fig:pipeline}(c). These adjusted features are mapped into the frozen emotion embedding space via the fusion encoder $E_{fuse}$. Formally, the emotion feature adjustment for frame $t$ can be denoted as follows:

\begin{equation}
    \tilde{f}^{label}_t=\frac{\hat{I}_t}{\left \| f^{label} \right \| } f^{label}
\end{equation}

During training, the facial animation is generated by combining content features from the ACM module with the emotion embedding from the FIM module. The total loss function comprises three components: reconstruction loss $\mathcal{L}_{recon}$, similarity loss $\mathcal{L}_{sim}$ between the prediction and the original emotion embedding, and intensity prediction loss $\mathcal{L}_{int}$.

\begin{equation}
    \mathcal{L}_{sim}=1-cos(E_e(\bm{R}_{1:T}), E_{fuse}(\tilde{\bm{F}}^{label}_{1:T}))
\end{equation}

\begin{equation}
    \mathcal{L}_{int}=\Vert Int(\hat{R}_{1:T})-Int(R_{1:T}) \Vert_2
\end{equation}

\begin{equation}
    \mathcal{L}_{FIM}=\mathcal{L}_{recon}+\lambda_{sim}\mathcal{L}_{sim}+\lambda_{int}\mathcal{L}_{int}
\end{equation}

By incorporating user-controllable emotion labels along with audio and text features, our approach enables precise generation of facial animations with dynamic emotional expressions.

\subsection{Multimodal Guided Animation} \label{sec:multi-modal}

While introducing dynamic emotional intensity allows for emotional fluctuations in facial animations, the overall expressiveness remains constrained by the predefined set of seven emotion labels. To enable fully user-controllable facial emotion features, we incorporate text descriptions and reference facial images as optional guidance, overcoming the limitations of predefined emotion labels.

\paragraph{\textbf{Data Collection}}
To guide expression generation, we collect reference images and corresponding text descriptions from a 2D talking head dataset, RAVDESS \cite{livingstone2018ryerson}. Specifically, Mediapipe \cite{lugaresi2019mediapipe} is employed to extract blendshape coefficients for each frame. Then, we select the frame with the highest emotional intensity as the reference image. To obtain a text description of the facial expression, we employ Google Gemini 2.0 Flash Thinking \cite{GoogleGemini}, which provides a natural language interpretation of the selected frame.

\paragraph{\textbf{Training}} 
Given CLIP’s \cite{radford2021learning} robust text and image encoding capabilities, we freeze the CLIP text encoder $E_{text}$ and image encoder $E_{image}$. To align these features with the emotion label embedding space, we introduce projection networks $P_{text}$ and $P_{image}$ to map the encoded features to latent features $f^{text}$ and $f^{image}$, which have the same dimension as $f^{label}$. In the multimodal guided generation mode, these multimodal embeddings replace the emotion label embeddings, while the rest of the model remains unchanged.

To maintain consistency between guided emotion features and motion-based emotion features, we train the model by freezing all components except $P_{text}$, $P_{image}$, and the fusion encoder $E_{fuse}$. The loss function includes reconstruction loss $\mathcal{L}_{recon}$ and similarity loss $\mathcal{L}_{sim}$, ensuring alignment between the guided emotion features and the generated motion, as shown below.

\begin{equation}
\begin{split}
    \mathcal{L}_{sim}=1-cos(E_e(\mathbf{R}_{1:T}),  E_{fuse}(\tilde{\mathbf{F}}^{g}_{1:T}))
\end{split}
\end{equation}

\begin{equation}
    \mathcal{L}_{CLIP}=\mathcal{L}_{recon}+\lambda_{sim}\mathcal{L}_{sim}
\end{equation}

where the emotion guidance modal $g \in \{ text, image \}$ can be either text or image.

By incorporating text descriptions and reference images, our method enables fully controllable expression generation. Combined with separate content and emotion embedding spaces, this approach produces vivid, audio-synchronized facial animations with dynamically adjustable and user-directed emotional expressions.

\section{Experiments}
\subsection{Experimental Settings}
\paragraph{\textbf{Datasets}}
Due to the limited availability of publicly accessible MetaHuman controller rig datasets, we utilize the dataset introduced by EmoFace \cite{liu2024emoface} for training audio-driven expression generation (Sec. \ref{sec:dynamic_emotion}). This dataset includes short-sentence utterances across seven emotional categories.

For multimodal guided animation (Sec. \ref{sec:multi-modal}), we employ RAVDESS \cite{livingstone2018ryerson}. Though EmoTalk \cite{peng2023emotalk} has extracted its blendshape coefficients for each frame, they cannot be directly used for training. We utilized Unreal Engine’s Live Link Plugin to convert the blendshape coefficients into MetaHuman controller values to bridge this gap.

\paragraph{\textbf{Implementation Details}}
For loss computation, we employ the Mean Square Error (MSE) to measure the difference between the predicted 174-dim rig parameters and ground truth values, defining the reconstruction loss $\mathcal{L}_{recon}$. The weights of other weight items are shared during training, with $\lambda_{sim}=0.1$, $\lambda_{int}=0.1$.

Unlike fully end-to-end training approaches, our method adopts a sequential training strategy, where the following components are trained separately: cross-reconstruction, audio content mapping, fusion intensity modeling, and multimodal guided generation. While in the cross-reconstruction stage, self-reconstruction, overlap exchange, and cycle exchange are also trained in order. All training phases utilize the Adam optimizer with a StepLR scheduler (drop rate $r=10$). The learning rate $\eta$ and decay rate $d$ differ for different training phases. In cross-reconstruction learning, $\eta=0.0001$, $d=0.995$, while $\eta=0.001$, $d=0.9$ in other training phases. As the model is designed to be very lightweight, the four training stages can be finished by a single NVIDIA RTX 3090 within 1 hour.

\paragraph{\textbf{Baseline Methods}}
We compare our MEDTalk with some publicly available audio-driven 3D talking-head animation models, namely FaceFormer \cite{fan2022faceformer}, EmoTalk \cite{peng2023emotalk}, EmoFace \cite{liu2024emoface}, DiffPoseTalk \cite{sun2024diffposetalk}. As these models were trained on different datasets, we adjusted the output dimension to 174 to match our dataset. We modified the style code to align with the embedding for different emotion labels. Each of the modified models was retrained on the same dataset until convergence and evaluated under identical test conditions.

\subsection{Quantitative Evaluation}
To evaluate the accuracy of generated animation, we propose metrics Mean Lip Error (MLE) and Mean Emotion Error (MEE) inspired by Lip Vertex Error and Emotion Vertex Error\cite{fan2022faceformer, richard2021meshtalk, peng2023emotalk}. We calculate the average $\ell_1$ error between the predicted and ground-truth lip and emotion region values.

Additionally, to evaluate the dynamic characteristics of facial expressions, we introduce Emotion Intensity Error (EIE) and Upper-Face Rig Deviation (FRD). EIE computes the $\ell_1$ error between the predicted and ground truth emotion intensity per frame (defined in Sec. \ref{sec:dynamic_emotion}). FRD \cite{pan2025vasa} quantifies the diversity of upper-face animations by comparing the average standard deviation between predicted and real animations. The evaluation metrics are defined as:

\begin{equation}
    \mathbf{MLE}(\hat{\bm{R}}, \bm{R})= \Vert \hat{\bm{R}}^{S_{lip}} - \bm{R}^{S_{lip}} \Vert_1
\end{equation}

\begin{equation}
    \mathbf{MEE}(\hat{\bm{R}}, \bm{R})= \Vert \hat{\bm{R}}^{S_{emo}} - \bm{R}^{S_{emo}} \Vert_1
\end{equation}

\begin{equation}
    \mathbf{EIE}(\hat{\bm{R}}, \bm{R})= \Vert Int(\hat{\bm{R}}) - Int(\bm{R}) \Vert_1
\end{equation}

\begin{equation}
    \mathbf{FRD}(\hat{\bm{R}}, \bm{R})=\frac{1}{\left\|S_{up}\right\|}\sum_{j \in S_{up}}(std(\bm{R}^j)-std(\hat{\bm{R}}^j))
\end{equation}

where $\bm{R}$ and $\hat{\bm{R}}$ represents ground truth and predicted controller rig sequences. $S_{lip}, S_{emo}, S_{up}$ are sets of controllers selected to represent lip movement, emotion, and upper face. $std(\cdot)$ represents the calculation of standard deviation.

\begin{table}
  \caption{Quantitative evaluation results. Best performance in \textbf{bold}, and the second best underlined.}
  \label{tab:sota}
  \begin{tabular}{lcccc}
    \toprule
    Methods & \textbf{MLE}$\downarrow$ & \textbf{MEE}$\downarrow$ & \textbf{EIE}$\downarrow$ & \textbf{FRD}$\downarrow$\\
    \midrule
    FaceFormer & 0.00662 & \underline{0.00952} & \textbf{0.69221} & 0.00618\\
    EmoTalk & 0.00756 & 0.02303 & 0.92046 & \underline{0.00116}\\
    EmoFace & \underline{0.00651} & 0.02072 & 0.88574 & 0.00507\\
    DiffPoseTalk & 0.01525 & 0.01664 & 0.89151 & \textbf{0.00075}\\
    \textbf{Ours} & \textbf{0.00596} & \textbf{0.00906} & \underline{0.79055} & 0.00289\\
  \bottomrule
\end{tabular}
\end{table}

The quantitative evaluation of MEDTalk and baseline models is presented in Tab. \ref{tab:sota}. The results indicate that MEDTalk consistently outperforms competing methods. MLE and MEE results demonstrate superior accuracy in lip shape and upper-face expressions, indicating that our disentangled content-emotion embedding space enables high-quality expressive facial animation with a lightweight model. EIE results show that MEDTalk achieves the second-best performance in accurately capturing emotional intensity dynamics, following FaceFormer. FRD results indicate that while MEDTalk slightly underperforms DiffPoseTalk (benefits from the diffusion model) and EmoTalk (incorporates a post-processing refinement step), it surpasses other baseline methods in expression diversity. Overall, the quantitative evaluation demonstrates that MEDTalk achieves state-of-the-art performance in terms of accuracy, expressiveness, and diversity of facial animation generation.

\subsection{Qualitative Evaluation}
\paragraph{\textbf{Disentangled Embedding Space}}

\begin{figure}[t]
  \centering
  \includegraphics[width=\linewidth]{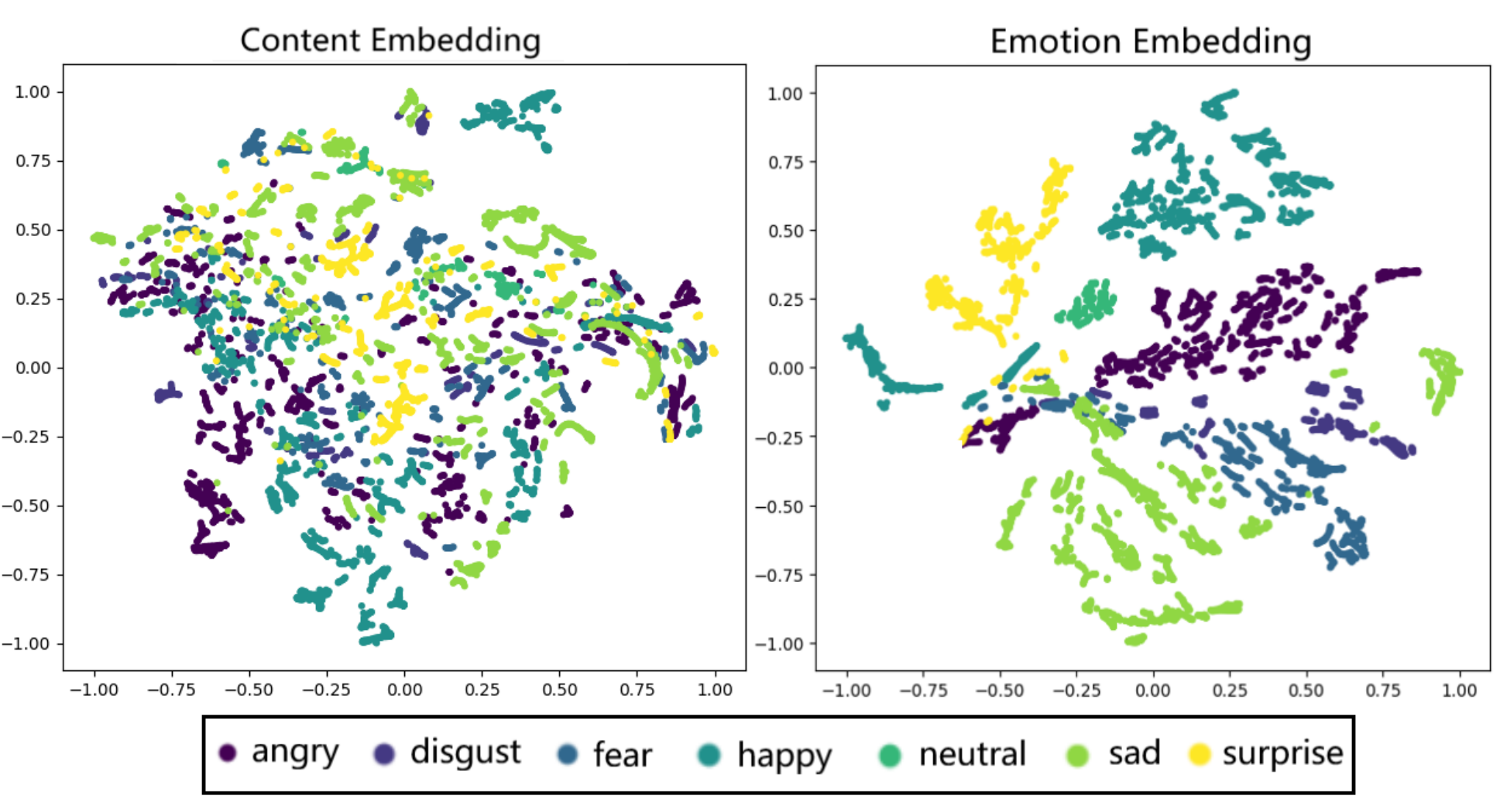}
  \caption{Disentangled Embedding Spaces.}
  \label{fig:embedding_space}
\end{figure}

\begin{figure}[t]
  \centering
  \includegraphics[width=0.85\linewidth]{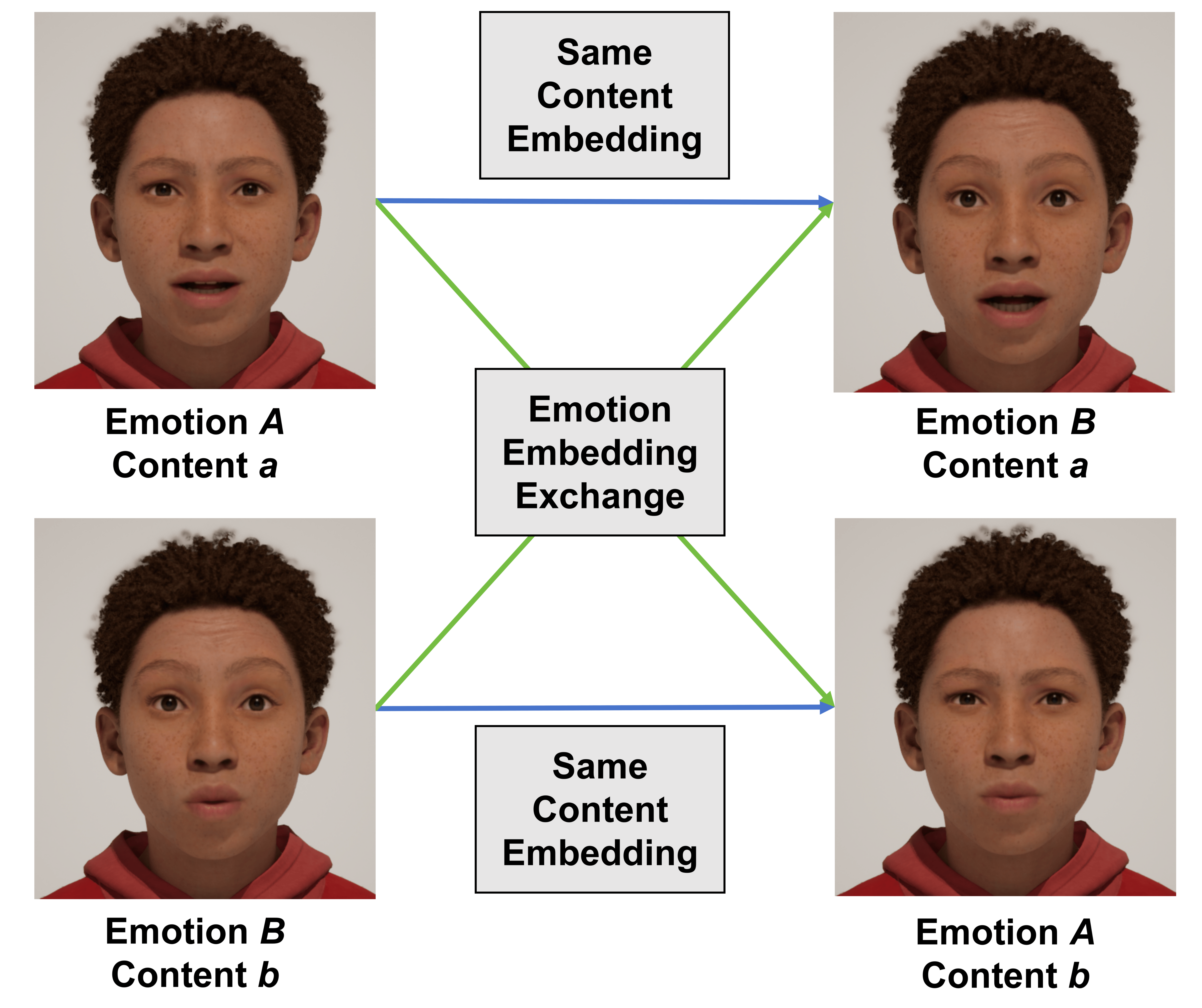}
  \caption{Swapping emotion embedding alters the emotional expression, with the lip shapes unaffected.}
  \label{fig:exchange_emotion}
\end{figure}

We employ T-SNE to visualize the content and emotion embedding spaces based on emotional labels, as shown in Fig. \ref{fig:embedding_space}. It demonstrates that distinct emotional categories are significantly separated in the emotion embedding space, whereas they remain indistinguishable in the content embedding space. Furthermore, we conduct cross-reconstruction by swapping the emotion embeddings of two entirely unrelated motion samples. As shown in Fig. \ref{fig:exchange_emotion}, the results indicate that modifying the emotion embeddings effectively alters the generated facial expressions while exerting minimal influence on lip shape.

\paragraph{\textbf{Visual Comparison}}

\begin{figure*}[t]
  \centering
 \includegraphics[width=0.9\linewidth]{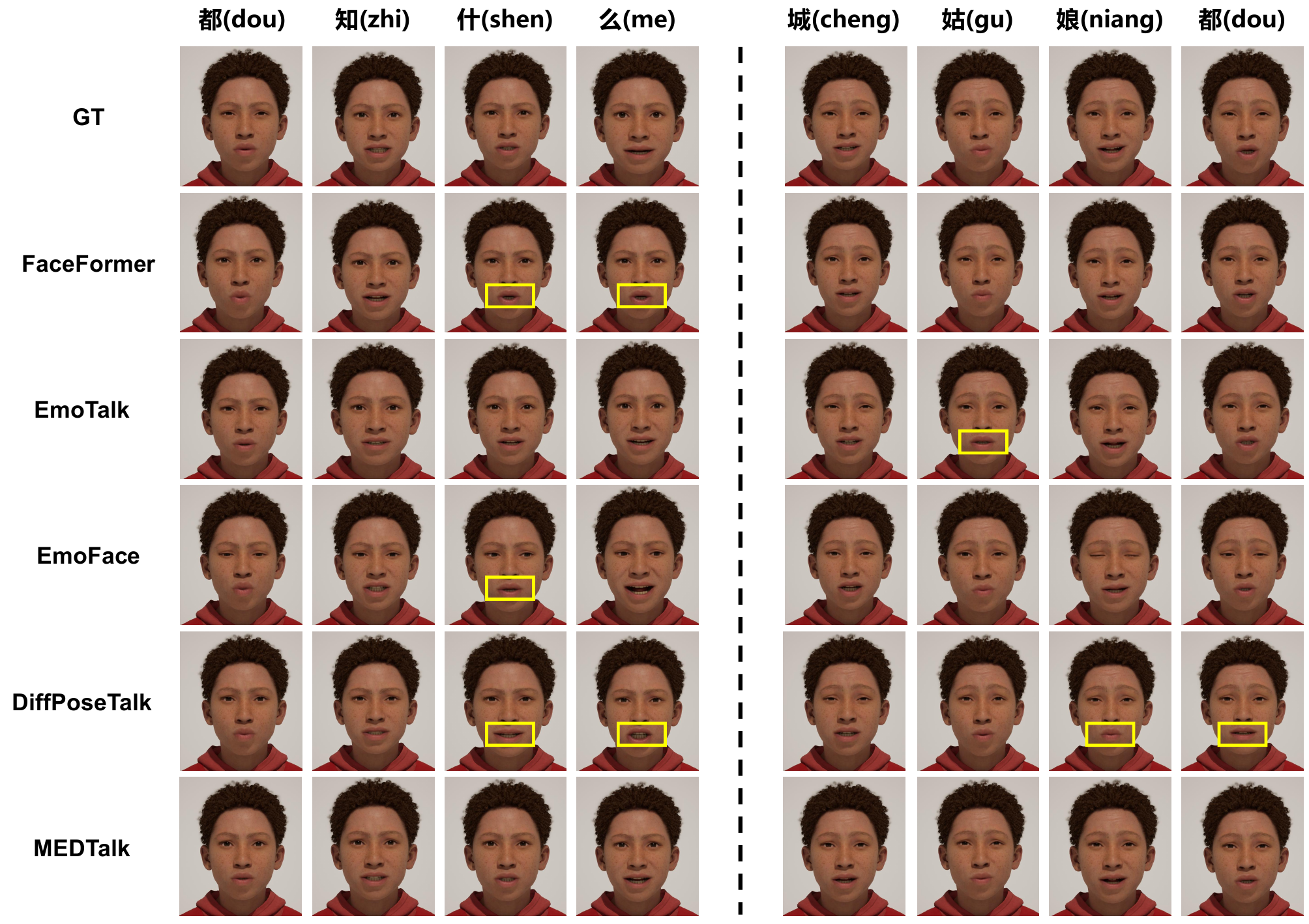}
    \caption{Qualitative comparison of facial movements generated by different methods under the "angry" (left) and "disgusted" (right) emotions, with annotated spoken words.}
    \label{fig:visual_comparison}
\end{figure*}

We compare MEDTalk with baseline models under the "angry" and "disgust" labels across different spoken utterances. The results presented in Fig. \ref{fig:visual_comparison} demonstrate that, while most methods generate natural lip movements, inconsistencies with the corresponding audio input may occur. In particular, except for MEDTalk, the lip shapes produced by other methods exhibit discrepancies, highlighting the advantages of our disentangled content and emotion embedding spaces. It enables independent control over lip shape and facial expressions, thereby mitigating mutual interference and enhancing overall accuracy.

\paragraph{\textbf{Multimodal Guided Generation}}

\begin{figure}[t]
  \centering
  \includegraphics[width=\linewidth]{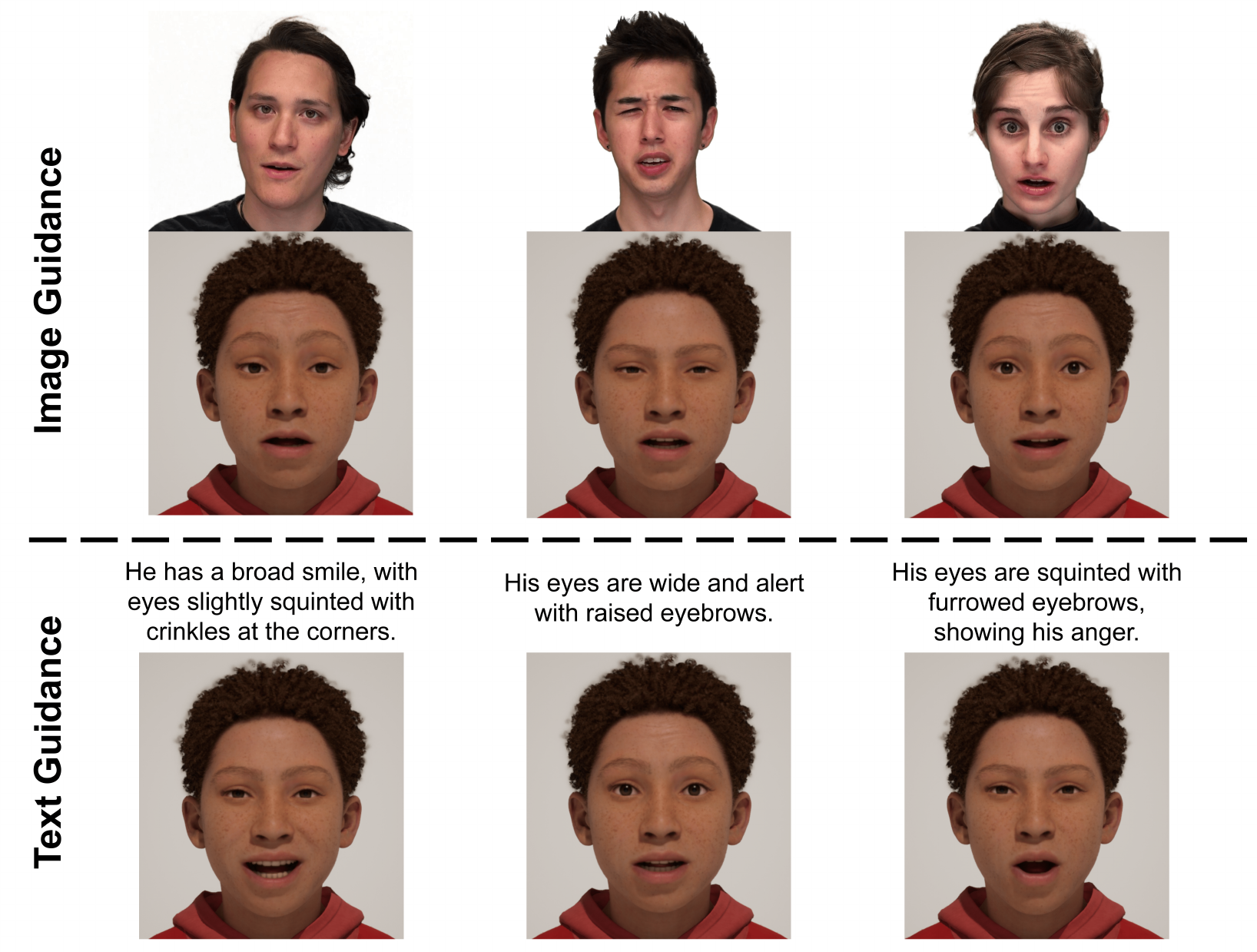}
  \caption{Results of multimodal guidance. Given reference images or text descriptions, MEDTalk generates facial animations that accurately reflect the specified speaking style.}
  \label{fig:multi_modal}
\end{figure}

To further evaluate MEDTalk, we conduct multimodal guided generation in the RAVDESS test set. As illustrated in Fig. \ref{fig:multi_modal}, the results generated exhibit a strong alignment with the inputs provided, achieving accurate and controllable emotional expression. Given that existing works adopt similar approaches for integrating multimodal guidance into talking head animation, our analysis focuses specifically on the performance of MEDTalk in this context.

\subsection{User Study}
We recruited a total of 42 participants with an average age of 20.5 years, ranging from 18 to 26 years old; 27 were male. All participants were unaware of the study objectives. A total of 10 audio clips from the test set were randomly selected for evaluation. Using the five methods and the ground truth, we generated animations for each clip, resulting in 60 video samples. The clips were presented to each participant in a randomized order. The participants rated each clip according to the accuracy of lip synchronization, emotional expressiveness, and overall facial animation quality. Ratings were given on a five-point Likert scale (1 = very poor, 5 = very good). The detailed results, summarized in Tab. \ref{tab:user_study}, indicate that MEDTalk outperforms other methods in all three metrics. This demonstrates the effectiveness of our proposed framework in generating more realistic and expressive facial animations.

\begin{table}
  \caption{Average rating with 95\% confidence interval}
  \label{tab:user_study}
  \begin{tabular}{lccc}
    \toprule
    Methods & \textbf{Lip-sync} & \textbf{Emotion Acc} & \textbf{Vividness}\\
    \midrule
    FaceFormer & \(3.262\pm0.094\) & \(3.970\pm0.072\) & \(2.996\pm0.242\) \\
    EmoTalk  & \(3.627\pm0.147\) & \(3.132\pm0.107\) & \(3.352\pm0.187\) \\
    EmoFace & \(3.886\pm0.112\) & \(4.083\pm0.068\) & \(3.416\pm0.198\) \\
    DiffPoseTalk & \(3.246\pm0.226\) & \(3.683\pm0.086\) & \(3.698\pm0.225\) \\
    \textbf{Ours} & \(4.022\pm0.137\) & \(4.132\pm0.093\) & \(3.762\pm0.184\) \\
    GT & \(4.421\pm0.072\) & \(4.390\pm0.057\) & \(4.123\pm0.147\) \\
  \bottomrule
\end{tabular}
\end{table}

\subsection{Ablation Study}

We further conduct ablation experiments to assess the effectiveness of each introduced component. The evaluation results are presented in Tab. \ref{tab:ablation}. In general, the experiment settings and corresponding analyses are summarized as:
\textbf{(a) w/o overlap:} We exclude the overlap exchange in the disentanglement training. As a result, the accuracy of the generated expression animation has dropped significantly, indicating that overlap exchange plays an important role in the decoupling process.
\textbf{(b) w/o cycle:} We exclude the cycle exchange in the disentanglement training. In this case, the accuracy of generated expressions also decreases, but smaller than (a). This indicates that the cycle exchange serves more as a regularization mechanism for models that already exhibit disentangling capability
\textbf{(c) w/o disentangle:} We train the model using audio-rig pairs in an end-to-end manner, which results in substantial increases in MLE, EIE, and FRD.
\textbf{(d) w/o intensity:} We use static emotion label embedding across the entire audio clip, which results in reduced accuracy and vividness in upper-face animations
\textbf{(e) w/o text:} We omit the textual input when predicting the intensity of emotion, which leads to noticeable deterioration, underscoring the necessity of incorporating both audio and text for dynamic emotion modeling

\begin{table}
  \caption{Ablation study for key components}
  \label{tab:ablation}
  \begin{tabular}{lcccc}
    \toprule
    Methods & \textbf{MLE}$\downarrow$ & \textbf{MEE}$\downarrow$ & \textbf{EIE}$\downarrow$ & \textbf{FRD}$\downarrow$\\
    \midrule
    w/o overlap & 0.01026 & 0.01112 & 0.90941 & \textbf{0.00134}\\
    w/o cycle & 0.00704 & 0.00934 & 0.82254 & 0.00832\\
    w/o disentangle & 0.00628 & \textbf{0.00861} & 0.83728 & 0.00825\\
    w/o intensity & 0.00665 & 0.01142 & 0.86488 & 0.00753\\
    w/o text & 0.00630 & 0.00965 & 0.83899 & 0.00757\\
    \textbf{Full model} & \textbf{0.00596} & 0.00906 & \textbf{0.79055} & 0.00289\\
  \bottomrule
\end{tabular}
\end{table}

\section{Discussion}
Despite the effectiveness of MEDTalk in generating vivid and expressive facial animations, certain limitations remain. First, after freezing the disentangled content and emotion latent spaces during the initial training stage, the generalization of lip shape generation is constrained by the diversity of the training data. This limitation becomes particularly evident in multilingual scenarios, where incorrect lip shapes may occur. Future work will focus on expanding the dataset with more diverse linguistic samples to enhance the generalizability of the disentangled representations. 
Moreover, while our current multimodal guidance mechanism effectively controls emotional embeddings, it does not extend to individual facial movements such as smiling or yawning. Future research will explore more comprehensive multimodal control strategies.

Leveraging the multi-character adaptation capabilities of Unreal Engine's MetaHuman, the rig parameters generated by MEDTalk can be seamlessly integrated into various avatars without additional adjustments. In fields such as virtual character animation and film production, our approach facilitates rapid generation of audio-synchronized lip movements. Furthermore, through multimodal expression control, our method allows animators and content creators to generate expressive animations by providing simple textual descriptions or reference images.

\section{Conclusion}

In this work, we introduce MEDTalk, a novel audio-driven 3D facial animation framework that enables precise, independent control over lip movements and facial expressions by disentangling content and emotion embeddings. Our approach enhances expressiveness by dynamically extracting emotion intensity from both audio and spoken text, allowing for more fluid and natural emotional transitions. Instead of relying on predefined emotion labels, MEDTalk supports multimodal input, enabling users to guide emotions through textual or visual cues. Our method captures subtle micro-expressions, enriching the realism and vividness of facial animations. Extensive experiments demonstrate that MEDTalk surpasses baseline models in lip synchronization, emotional expressiveness, and overall animation quality, offering a versatile and effective solution for controllable 3D talking head animation.

\section{Acknowledgement}
This work is supported by National Natural Science Foundation of China (NSFC, No. 62472285 and No. 62102255), Sponsored by CCF-NetEase ThunderFire Innovation Research Funding (NO. CCF-Netease 202508 and NO. CCF-Netease 202509).
\bibliographystyle{ACM-Reference-Format}
\bibliography{reference}

\end{document}